\pdfoutput=1

\documentclass[11pt]{article}

\usepackage{EMNLP2023}

\usepackage{times}
\usepackage{latexsym}

\usepackage[T1]{fontenc}

\usepackage[utf8]{inputenc}

\usepackage{microtype}

\usepackage{longtable} 
\usepackage{geometry} 
\usepackage{inconsolata}
\usepackage{graphicx}
\usepackage{booktabs}
\usepackage{multirow}
\usepackage{array}
\usepackage{amsmath}
\usepackage{enumitem}
\usepackage{listings}
\usepackage{xcolor}
\usepackage{hyperref}

\title{Self-Reflective Planning with Knowledge Graphs:\\Enhancing LLM Reasoning Reliability for Question Answering}

\author{Jiajun Zhu, Ye Liu, Meikai Bao, Kai Zhang, Yanghai Zhang, Qi Liu \\ 
  State Key Laboratory of Cognitive Intelligence, University of Science and Technology of China
    \\\texttt{\{jiajunzhu, baomeikai, apocalypseh\}@mail.ustc.edu.cn}, 
    \\\texttt{yeliu.liuyeah@gmail.com}, 
    \\\texttt{\{kkzhang08, qiliuql\}@ustc.edu.cn}
}

\begin{document}
\maketitle

\begin{abstract}
Recently, large language models (LLMs) have demonstrated remarkable capabilities in natural language processing tasks, yet they remain prone to hallucinations when reasoning with insufficient internal knowledge. While integrating LLMs with knowledge graphs (KGs) provides access to structured, verifiable information, existing approaches often generate incomplete or factually inconsistent reasoning paths. To this end, we propose Self-Reflective Planning (SRP), a framework that synergizes LLMs with KGs through iterative, reference-guided reasoning. Specifically, given a question and topic entities, SRP first searches for references to guide planning and reflection. In the planning process, it checks initial relations and generates a reasoning path. After retrieving knowledge from KGs through a reasoning path, it implements iterative reflection by judging the retrieval result and editing the reasoning path until the answer is correctly retrieved. Extensive experiments on three public datasets demonstrate that SRP surpasses various strong baselines and further underscore its reliable reasoning ability. 

\end{abstract}

\section{Introduction}

Recent breakthroughs in large language models (LLMs) have significantly advanced natural language processing \citep{brown2020language, ouyang2022training, bubeck2023sparks}. Nonetheless, LLMs largely depend on knowledge gained through pretraining, and consequently, they tend to struggle with hallucination when solving questions that extend beyond their scope of knowledge. 

To tackle these problems, researchers attempt to integrate LLMs with knowledge graphs (KGs), capitalizing on the explicit structure and semantic knowledge to reinforce factual grounding \citep{hogan2021knowledge}. Prior research has explored two categories of approaches: (1) fine-tuning methods that equip models with the ability to utilize information in KGs \citep{shi-etal-2021-transfernet,zhang-etal-2022-subgraph}; and (2) guiding LLMs with prompts to make plans and decisions in order to search for the knowledge in KGs for answering questions \citep{li-etal-2023-shot,cheng2024call}. 

\begin{figure}[t]
    \centering
    \vspace{0.5cm}
    \includegraphics[width=1\linewidth]{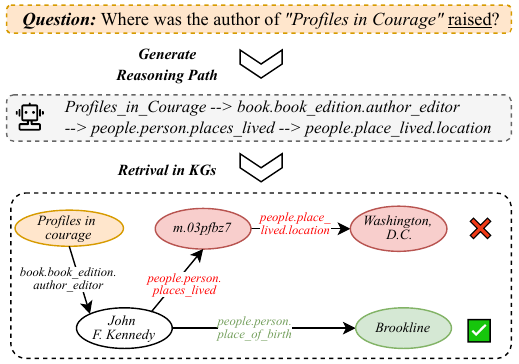}
    \caption{An illustration of the question answering over knowledge graphs.}
    \label{fig:intro}
\end{figure}

Despite their effectiveness, existing approaches still face notable challenges. On the one hand, fine-tuning-based models require high-quality, large-scale data for fine-tuning. Such requirements not only limit their adaptability to new domains but also demand substantial computational resources. On the other hand, prompt-based approaches primarily focus on planning effective retrieval, while falling short in reflecting the retrieved information. For example, in Figure~\ref{fig:intro}, given the question ``\textit{Where was the author of `Profiles in Courage' raised}'', the model generates a reasoning path to search for relevant knowledge in the KG. The result obtained from the reasoning path is ``\textit{Washington, D.C.}'', which is incorrect as the relation ``\textit{people.place\_lived.location}'' in the reasoning path points to where the person lived, but the question asks about where the person ``was raised''. While the model retrieves a relevant result, it lacks reflection on whether the reasoning path truly aligns with the intent of the question. As a consequence, it overlooks the correct relation ``\textit{people.person\allowbreak.place\_of\_birth}'' and correct entity ``\textit{Brookline}''.


To address these shortcomings, we propose Self-Reflective Planning (SRP), a KG-enhanced LLM framework with reliable planning and reflection ability. Specifically, SRP first designs a Reference Searching module, which searches for relevant references as evidence. Subsequently, a Path Planning module is designed to check initial relations linked to topic entities and generate the reasoning path. After that, we develop a Knowledge Retrieval module to instantiate the reasoning path with the KG. More importantly, to achieve reliable reflection, a Reflection and Reasoning module is employed to iteratively judge retrieval results and edit the reasoning path until the answer is retrieved.


In summary, the main contributions of our work could be summarized as follows.
\begin{itemize}[leftmargin=12pt, itemsep=0pt, parsep=3pt, topsep=5pt, partopsep=5pt]
\item We explore methods for reliable planning and reflection with LLMs to generate accurate reasoning path, offering a new perspective for building more reliable question answering algorithms.
\item We propose a Self-Reflective Planning (SRP) framework, which can generate the effective reasoning path and refine reasoning errors with reliable reflection.
\item Extensive evaluations on WebQSP, CWQ, and GrailQA datasets demonstrate SRP’s superior performance compared to competitive baselines, underscoring the reliability and accuracy of SRP. The code is available on \href{https://anonymous.4open.science/r/SRP-E06C}{https://anonymous.4open.science/r/SRP-E06C}.
\end{itemize}
\section{Related Work}


\paragraph{LLM Reasoning.} To promote reasoning ability of LLMs, numerous researchers have directed LLMs to incorporate their thought processes into their outputs rather than merely delivering direct answers \citep{wei2022chain, hao2024training, kojima2022large}. Initially, Chain of Thought (CoT) \citep{wei2022chain} framework was developed to present several examples of intermediate reasoning steps in natural language as prompts. Subsequently, various adaptations of CoT reasoning, such as Self-Consistency \citep{wang2022self}, Tree-of-Thought \citep{yao2023tree}, Graph-of-Thought \citep{besta2024graph}, were introduced to motivate the reasoning ability of LLMs. To facilitate LLMs in achieving a cognitive process that parallels human thinking, many studies \citep{madaan2023self, wang2024theoretical, guan2024amor, shinn2023reflexion} have devised self-correction mechanisms using feedback to amend faulty reasoning and ensure precision. Our work is also inspired by self-correction and proposes a self-reflection mechanism to better plan and correct the reasoning path for searching effective knowledge in the KG. 

\begin{figure*}[ht]
    \centering
    \includegraphics[width=1\linewidth]{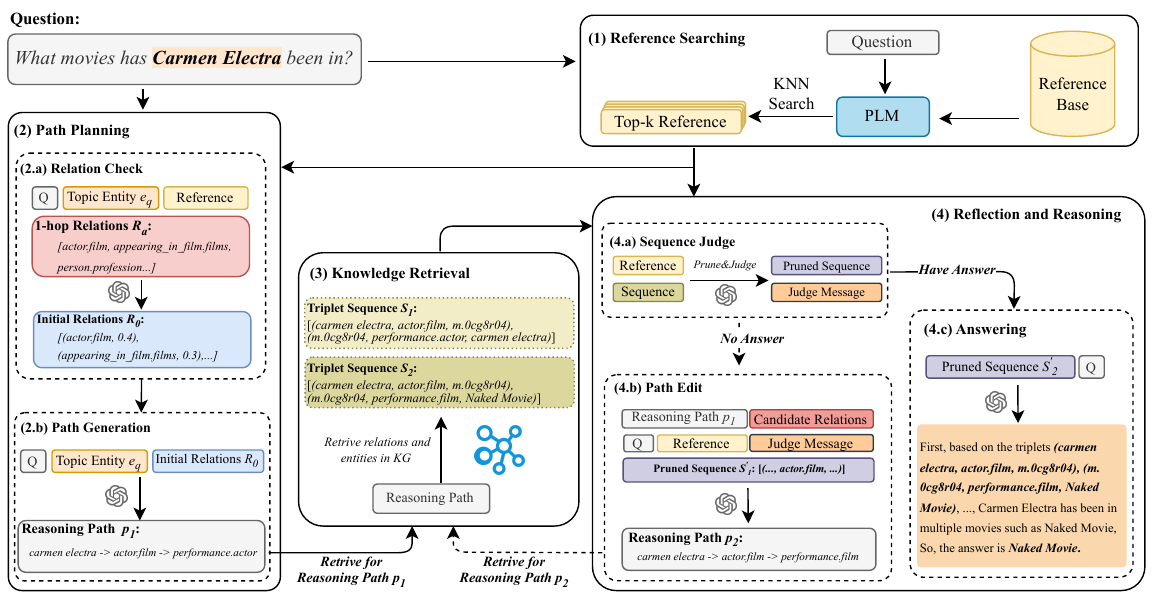}
    \caption{The framework of our SRP method. It consists of four main parts: (1) Reference Searching, (2) Path Planning, (3) Knowledge Retrieval and (4) Reflection and Reasoning.}
    \label{fig:model}
\end{figure*}

\paragraph{KG-enhanced LLM.} LLMs are pre-trained on extensive datasets, but they still encounter issues such as outdated information, hallucinations, and unclear decision-making processes \citep{zhang2024pretraining,zhang2023siren,zhang2024debiasing}. A promising solution to these problems is utilizing KGs to provide explicit and modifiable knowledge to LLMs. Earlier research incorporated KGs during the pre-training \citep{wang2021kepler} or fine-tuning \citep{luo2023reasoning} phases, but these methods demand substantial computational resources and perform poorly when dealing with problem that were not encountered during the training or fine-tuning period. Consequently, some approaches \citep{baek2023knowledge,yang2024give,cheng2024call} first extracted information from KGs and then directly supplied explicit knowledge to LLMs. These methods eliminate the requirement for pre-training or fine-tuning the model and achieve satisfactory performance. However, this method still falls short in accurately extracting question-relevant knowledge from the KG, which reduces the reliability of model. To solve this problem, our work can improve reliability of reasoning path through reference searching, checking the first relation and self-reflection, which is effective for retrieve relevant knowledge in KGs.

\section{Problem Definition}

Combining KGs with LLMs to answer questions belongs to knowledge graph question answering (KGQA) task, which is about resolving natural language question with KGs. KG is a set of triples, i.e., $\{(e, r, e'){\mid}e, e' \in \mathcal{E}, r \in \mathcal{R}\}$, where $\mathcal{E}$ and $\mathcal{R}$ are the sets of entities and relations. Given a question $q$, a set of topic entities $\mathcal{T}_q$ extracted within $q$, and a KG $\mathcal{G}$, the objective of KGQA is to infer valid answers $\mathcal{A}_q$. To achieve this, LLM needs to plan reliable reasoning paths, retrieve the triplet sequences based on reasoning paths, and answer questions with triplet sequence. The reasoning path and triplet sequence are defined as below.


\paragraph{Reasoning Path.} Reasoning Path is sequences starting from entity and followed by one or more relations, e.g., $p = e_0 \to \hat{r}_0 \to \hat{r}_1 \to \dots \to \hat{r}_l$, where each $\hat{r}_i$ is a predicted relation generated by LLM and $l$ is the length of reasoning path.

\paragraph{Triplet Sequence.} Triplet Sequence is a sequence of triplets retrieved from KGs. Given reasoning path $p$, the triplet sequence is $S = \{(e_{d},~r_{d},~e_{d+1})\}^D_{d=0}$, where $e_d, e_{d+1} \in \mathcal{E}$ and $r_d \in \mathcal{R}$. Noted that each $r_d$ is corresponding to a predicted relation $\hat{r}_d$ in $p$.

\section{Methodology}

\subsection{Overview}
As shown in Figure~\ref{fig:model}, our SRP includes four modules: 1) Reference Searching module that searches for reference, 2) Path Planning module that generates reasoning path, 3) Knowledge Retrieval module that retrieves triplet sequence and 4) Reflection and Reasoning module that performs judgement, editing and answering.

Given a question, module 1) first searches for references to guide module 2) and 4). Then, module 2) generates initial reasoning path. Module 3) bridges Module 2) and Module 4), achieving iterative retrieval and refinement of reasoning paths. The prompts of SRP are illustrated in Appendix~\ref{appendix:prompts}.

\subsection{Reference Searching}
\label{subsec:reference_searching}
To guide the reasoning process, module retrieves relevant cases from reference base, which contains massive referable cases about solving question with KGs. A case includes a question, reasoning paths that can be utilized to retrieve answers in KGs and the correct answers of the question, which is denoted as reference. We extract references from training set. To construct the reference base, each question in training set is encoded into dense vectors with pretrained language model and then clustered (e.g., via K-Means). We randomly select the same number of references from each cluster and store them in the reference base.  

During process of planning and self-reflection, the model encodes the input question and compares it against the embeddings of questions from reference base using similarity-based search (e.g., K-Nearest Neighbors) to retrieve the Top-k most similar corresponding references. These references are integrated into downstream modules to enrich the reasoning process with relevant historical contexts, as illustrated in Figure~\ref{fig:model}, thereby motivating reasoning ability of LLM to enhance accuracy and reliability of planning and self-reflection.

\subsection{Path Planning}
\label{subsec:initial_plan}

The Path Planning module is designed to generate reliable initial reasoning path, which includes two parts: relation check and path generation. 

\paragraph{Relation Check.} Before path generation, it is necessary to check each 1-hop relation of topic entity $e_{q}$, which means relation directly connected to $e_{q}$. This is because the LLM lacks context knowledge of $e_{q}$ in KGs. Checking the 1-hop relations connected to $e_{q}$ before generating the reasoning path $p$ can help LLMs predict reliable $p$ and align $p$ with the actual structure of the KG, thereby enhancing the reliability of reasoning path. It identifies the most relevant 1-hop relations connecting $e_{q}$ to other entities within the KG. The 1-hop relations of $e_{q}$ are denoted as $R_{a} = \{r^a_{0}, r^a_{1}, r^a_{2} \dots\}$. Guided by the information from the references obtained in reference searching, the LLM assesses each relation in $R_a$ based on its relevance to the given question $q$. The LLM scores the $R_a$, and the top-K relations with the highest scores are selected as the initial relations $R_0 = \{r^0_{0}, r^0_{1}, \dots, r^0_{k}\}$ to participate in the path generation.   

For demonstration, consider the example in Figure \ref{fig:model}, the LLM checks 1-hop relations of $e_{q}$. The LLM scores 1-hop relations of \textit{carmen electra}, such as \textit{actor.film}, \textit{appearing\_in\_film.film}, \textit{person.profession} and so on. The relevance scores assigned by the LLM are used to select relations including \textit{actor.film} and \textit{appearing\_in\_film.film} as the top-K candidate relations, with indicative scores of 0.4 and 0.3, respectively. These selected $R_0$ serve as candidates for the first relation starting from $e_q$ in reasoning path, thereby reducing uncertainty in the subsequent path generation.

\paragraph{Path Generation.} The path generation process generates the complete reasoning path $p$ starting from the topic entity $e_q$ with LLM, leveraging the initial relations $R_0$ obtained through relation check. Continuing the example of Figure \ref{fig:model}, the LLM predicts $p$ starting from the topic entity \textit{carmen electra}. Taking inputs that include the question $q$, $R_0$ (e.g., \textit{actor.film}), and $e_q$, the LLM generates the reasoning path ``\textit{carmen electra $\rightarrow$ actor.film $\rightarrow$ performance.actor}'', denoted as the reasoning path $p_1$. This predicted reasoning path serves for further retrieval in KGs and might be refined one or more times to improve reliability of the reasoning path.

\subsection{Knowledge Retrieval}

The Knowledge Retrieval module aims to instantiate the reasoning path $p = e_0 \to \hat{r}_0 \to \hat{r}_1 \to \dots \to \hat{r}_l$ by searching the corresponding triplet sequence $S = \{(e_{d},~r_{d},~e_{d+1})\}^D_{d=0}$, where $D$ is the length of $s$, and $D \leq l$. Each $r_d$ in $S$ has a corresponding predicted relations $\hat{r}_d$ in $p$. 

Following method in Readi \citep{cheng2024call}, we conduct retrieval by comparing semantic similarity. The retrieval is starting from $e_0$, which is also the topic entity $e_q$ of the question. To retrieve relation corresponding to $\hat{r}_0$, we first utilize BM25 and Contriever \citep{izacard2022unsupervised} to obtain relations semantically similar to $\hat{r}_0$, denoted as $R^s_0 = \{r^s_{0,~0},~r^s_{0,~1},~\dots\}$. Then, we examine the 1-hop relations connected to $e_0$. If there is one or more relations among $e_0$ that exists in $R^s_0$, then the corresponding relation $r_0$ and entity $e_1$ in KGs will be retrieved. The retrieval process will be iterated until each predicted relation in $p$ finds corresponding relation in KGs or a predicted relation in $p$ fails to find corresponding relation in KGs.  

For the instance in Figure \ref{fig:model}, the reasoning path $p_1$ from Path Planning is successfully retrieved, whose corresponding triplet sequence is \textit{[(carmen electra, actor.film, m.0cg8r04), (m.0cg8r04, performance.actor, carmen electra)]}, denoted as triplet sequence $S_1$. The edited path generated from path edit (reasoning path $p_2$) is also retrieved as \textit{[(carmen electra, actor.film, m.0cg8r04), (m.0cg8r04, performance.film, Naked Movie)]}, denoted as triplet sequence $S_2$.

\subsection{Reflection and Reasoning}
\label{subsec:self_reflection}

The Reflection and Reasoning module enables LLM to evaluate retrieved triplets, then choose to refine the reasoning path for next retrieval in KGs or answer question with retrieved information based on evaluation result. It consists of three parts: sequence judge, path edit and answering. 

\paragraph{Sequence Judge.} The sequence judge aims to determine whether the current reasoning path $p$ needs to be edited, and retain the subsequence related to the question in the triplet sequence $S$ to eliminate irrelevant triplets. This process is designed to provide reliable information for answering questions and path editing.  

Under the guidance of references, LLM assesses whether the answer of question or relevant information is in triplets of $S$ and outputs judgement result and its thinking process. If the judgement result is ``have answer'', $p$ will not be edited and self-reflection will be stopped, then SRP will answer the question. If the judgement result is ``no answer'', $p$ will be edited in path edit part for next retrieval. LLM will also generate pruned sequence $S^{\prime} = \{(e^{\prime}_{d},~r^{\prime}_{d},~e^{\prime}_{d+1})\}^{D^{\prime}}_{d=0}$, where $D^{\prime} \leq D$. It is the sub-sequence of $S$ which relevant to answering question. By evaluating $S^{\prime}$, we can identify that there might be predicted relations in $p$ that need to be corrected. These relations may not be retrievable in the KG or may not be helpful for answering the question. The evaluation result of $S^{\prime}$ and thinking process of LLM are concatenated as judge message for path edit process.

For the example in Figure~\ref{fig:model}, the triplet sequence $S_1$ is judged to have no answer, and the LLM considers \textit{(m.0cg8r04, performance.actor, carmen electra)} is meaningless for answering question, so LLM generated pruned sequence $S^{\prime}_1$ \textit{[(carmen electra, actor.film, m.0cg8r04)]} and corresponding judge message. For triplet sequence $S_2$, LLM judges that answer is in the sequence, so the self-reflection is stopped and the model will answer question with pruned sequence $S^{\prime}_2$ which is same as triplet sequence $S_2$.

\paragraph{Path Edit.} Path edit refines the reasoning path $p$ to better align with the question $q$ and reasoning goals, triggered when the triplet sequence $S$ lacks sufficient information. Path edit requires the 1-hop relations linked to the tail entity of the pruned sequence $S^{\prime}$ as the candidate relations for the edit process, which is obtained from retrieval module. As shown in Figure~\ref{fig:model}, with information from references, judge message and $S^{\prime}$, LLM edits the reasoning path $p_1$ with candidate relations. It modifies the relation \textit{performance.actor} to \textit{performance.film} and generates edited reasoning path ``\textit{carmen electra $\rightarrow$ actor.film $\rightarrow$  performance.film}'', denoted as reasoning path $p_2$.

\paragraph{Answering.} If the judgement result is ``have answer'', LLM will reason for answering question based on pruned sequence $S^{\prime}$ from sequence judge part. Using Chain-of-Thought (CoT) reasoning prompts, the LLM generates the final answer of question based on $S^{\prime}$ . This step involves analyzing the triplets in $S^{\prime}$ to deduce the answer. For the instance in Figure~\ref{fig:model}, based on triplets \textit{(carmen electra, actor.film, m.0cg8r04)}, \textit{(m.0cg8r04, performance.film, Naked Movie)}, LLM can give conclusion that ``\textit{the answer is Naked Movie}''.

\section{Experiments}

\subsection{Experiment Setup}
\paragraph{Datasets and Evaluation Metrics.} To evaluate the reasoning ability of SRP, we conduct experiments on three multi-hop KGQA datasets called WebQuestionsSP (WebQSP) \citep{yih-etal-2016-value}, ComplexWebQuestions (CWQ) \citep{talmor-berant-2018-web} and GrailQA \citep{gu2021beyond}. More statistics are illustrated in Table~\ref{tab:data}. Besides, all of the three datasets utilize Freebase \citep{bollacker2008freebase} as the background KG. To ensure computational efficiency in processing the large-scale GrailQA dataset, our experimental framework adopts the same test samples established in ToG \citep{sun2023think}. You can refer to Appendix~\ref{appendix:data} for more details.

\begin{table}[t]
  \centering
  \setlength{\tabcolsep}{8pt} 
    {
    \begin{tabular}{lccc}
    \toprule
    \textbf{Datasets} & \textbf{\#Train} & \textbf{\#Test} & \textbf{\#Max hop} \\
    \midrule
    WebQSP & 3,098  & 1,628  & 2 \\
    CWQ   & 27,639 & 3,531  & 4 \\
    GrailQA & 44,337 & 1,000  & 4 \\
    \bottomrule
    \end{tabular}}
  \caption{Statistics of three datasets.}
  \label{tab:data}%
\end{table}%

\begin{table*}[tp]
  \centering
  \setlength{\tabcolsep}{5pt}
    {
    \begin{tabular}{lcccccc}
    \toprule
    \multirow{2}[4]{*}{\textbf{Method}} & \multirow{2}[4]{*}{\textbf{WebQSP}} & \multirow{2}[4]{*}{\textbf{CWQ}} & \multicolumn{4}{c}{\textbf{GrailQA}} \\
\cmidrule{4-7}          &       &       & \textbf{overall} & \textbf{I.I.D.} & \textbf{Compositional} & \textbf{Zero-shot} \\
    \midrule
    \multicolumn{7}{c}{Fine-tuned Method} \\
    \midrule
    TransferNet$^{*}$ & 71.4  & 48.6  & -     & -     & -     & - \\
    TIARA$^{*}$ & 75.2  & -     & 73.0  & \textbf{87.8}  & \textbf{69.2}  & 68.0 \\
    SR+NSM+E2E$^{*}$ & 69.5  & 49.3  & -     & -     & -     & - \\
    Flexkbqa$^{*}$ & 60.6  & -     & 62.8  & 71.3  & 59.1  & 60.6 \\
    \midrule
    \multicolumn{7}{c}{LLM-only Method} \\
    \midrule
    GPT3.5 & 63.8  & 45.7  & 25.8  & 20.0  & 18.7  & 30.8 \\
    GPT-4.1-mini & 64.2  & 52.4  & 36.5  & 34.2  & 26.3  & 41.1 \\
    \midrule
    \multicolumn{7}{c}{Prompting LLM Methods} \\
    \midrule
    KB-BINDER$^{*}$ & 74.4  & -     & 50.6  & -     & -     & - \\
    StructGPT$^{*}$ & 72.6  & 54.3  & -     & -     & -     & - \\
    ToG-GPT3.5$^{*}$ & 76.2  & 57.1  & 68.7     & 70.1     & 56.1     & 72.7 \\
    Readi-GPT3.5 & 74.5  & 56.7  & 67.0  & 67.9  & 53.5  & 71.4 \\
    Readi-GPT4.1-mini & \underline{80.9}  & \underline{60.2}  & \underline{71.7}  & 67.5  & 60.1  & \underline{77.6} \\
    \midrule
    SRP-GPT3.5 (ours) & 78.6  & 58.7  & 71.2  & 68.3  & 58.6  & 76.9 \\
    SRP-GPT4.1-mini (ours) & \textbf{83.6} & \textbf{69.0} & \textbf{78.8} & \underline{75.8} & \underline{62.6} & \textbf{85.8} \\
    \bottomrule
    \end{tabular}}
  \caption{Performance comparison with different baselines on three KGQA dataets. \textbf{Bold font} represents the optimal result, while the \underline{underline font} represents the sub-optimal result. The results labeled with * are cited from their original publications.}
  \label{tab:table_2}%
\end{table*}%

Consistent with established methodologies in previous work \citep{cheng2024call, li2023few, jiang2023structgpt}, we employ exact match accuracy (Hits@1) as the primary evaluation metric for model performance assessment.

\paragraph{Implementation Details.} In Reference Searching and Knowledge Retrieval module, we adopt the all-MiniLM-L6-v2 based on sentence-transformers \citep{reimers-gurevych-2019-sentence} as the representation model. We extract 100 questions and corresponding references from each dataset to construct reference base. The number of retrieved reference for each test question is set to be k = 4. 

We adopt gpt-3.5-turbo \citep{openai_gpt35_turbo} and gpt-4.1-mini \citep{openai_gpt4.1_mini} as the LLMs to conduct experiments, denoted as SRP-GPT3.5 and SRP-GPT4.1-mini. Temperature is 0.3 for all modules. Following previous research \citep{cheng2024call}, we utilize a Pyserini as a hybrid searcher with BM25 and Contriver \citep{izacard2022unsupervised}. For each relation in reasoning path, we retrieve top-5 similar relations on Freebase. We deploy Virtuoso server to search for knowledge from Freebase. 

\paragraph{Benchmark Methods.} In order to verify the performance of our SRP model, we compare SRP with the state-of-the-art KGQA methods: 1) Fine-Tuned model methods, including TransferNet \citep{shi-etal-2021-transfernet}, TIARA \citep{shu-etal-2022-tiara}, SR+NSM+E2E \citep{zhang-etal-2022-subgraph} and Flexkbqa \citep{li2024flexkbqa}; 2) LLM-only methods, including GPT3.5 \citep{openai_gpt35_turbo} and GPT4.1-mini \citep{openai_gpt4.1_mini}; 3) Prompting LLM methods, including KB-BINDER \citep{li-etal-2023-shot}, StructGPT \citep{jiang2023structgpt}, ToG \citep{sun2023think} and Readi \citep{cheng2024call}. It is worth noting that, we utilize GPT3.5 and GPT4.1-mini to motivate Readi in experiment, denoted as Readi-GPT3.5 and Readi-GPT4.1-mini. Results of ToG (ToG-GPT3.5) are cited from \citep{chen2024plan}. More details can be seen from Appendix~\ref{appendix:baseline}.

\subsection{Experiment Result}

As shown in Table~\ref{tab:table_2}, SRP consistently demonstrates state-of-the-art (SOTA) performance on both the WebQSP and CWQ benchmarks, while achieving the second-highest overall performance on GrailQA, demonstrating its capacity to effectively leverage an external KG for robust retrieval and multi-hop reasoning. First, by integrating a systematic retrieval module with self-reflection reasoning prompts, SRP not only detects relevant facts more accurately but also navigates complex reasoning pathways in a manner that outperforms prompting-based LLM baselines such as Readi. This superiority becomes evident on both GPT-3.5 and GPT-4.1-mini backbones, highlighting SRP’s adaptability to different large language model architectures and underscoring the versatility of our retrieval-centric design. Second, although SRP does not incorporate any pre-training or fine-tuning steps, it still surpasses notable fine-tuned models, including TIARA, SR+NSM+E2E, and Flexkbqa; this observation underscores the power of a well-designed prompting approach to capture nuanced relationships within a knowledge graph, thus compensating for the absence of additional parameter updates. Third, when compared against LLM-only methods—where no external knowledge is used—SRP reveals significant gains in accuracy, indicating that the KG provides crucial information and helps avoid the hallucinations or logical gaps that can arise when relying purely on learned language representations.

\begin{figure*}[htp]
    \centering
    \includegraphics[width=0.9\linewidth]{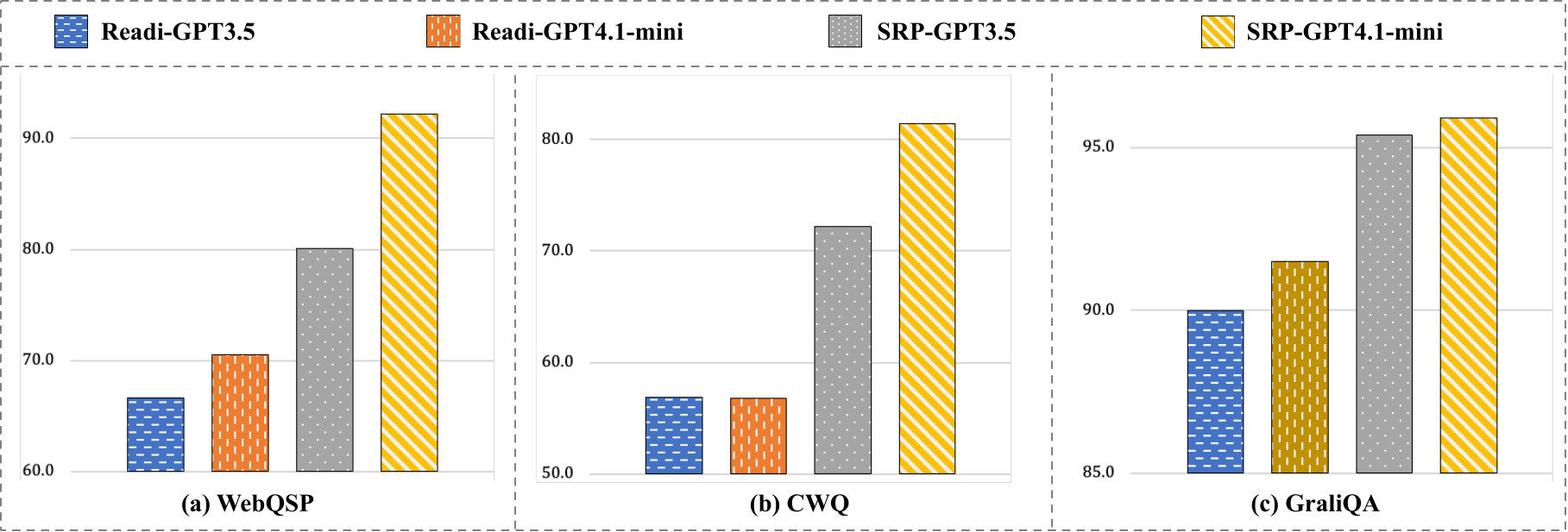}
    \caption{Reliable answering rate of Readi and SRP on theree datasets.}
    \label{fig:data}
\end{figure*}

\begin{table}[t]
  \centering
  \setlength{\tabcolsep}{3pt} 
    {
    \begin{tabular}{lccc}
    \toprule
    \textbf{Method} & \textbf{WebQSP} & \textbf{CWQ} & \textbf{GraliQA} \\
    \midrule
    \textbf{SRP} & \textbf{78.5} & \textbf{58.7} & \textbf{71.2} \\
    \midrule
    w/o relation check & 76.9  & 56.7  & 62.9 \\
    w/o self-reflection & 76.9  & 55.5  & 68.7 \\
    w/o reference & 75.7  & 57.6  & 69.8 \\
    w/ random reference & 77.0  & 58.2  & 70.9 \\
    \bottomrule
    \end{tabular}%
  }
  \caption{Ablation study}
  \label{tab:table_3}%
\end{table}%

\begin{table}[tp]
  \centering
  \setlength{\tabcolsep}{2.5pt} 
    {
    \begin{tabular}{lccc}
    \toprule
    \textbf{Methods} & \textbf{WebQSP} & \textbf{CWQ}   & \textbf{GraliQA} \\
    \midrule
    Readi-GPT3.5 & 55.7  & 44.2  & 68.3 \\
    Readi-GPT4.1-mini & 64.3  & 41.6  & 74.2 \\
    SRP-GPT3.5 & 69.9  & 60.5  & 80.3 \\
    SRP-GPT4.1-mini & 85.3  & 70.3  & 85.0 \\
    \bottomrule
    \end{tabular}%
  }
  \caption{Searching success rate of Readi and SRP on theree datasets.}    
  \label{tab:table_4}%
\end{table}%

\subsection{Ablation Study}
Table~\ref{tab:table_3} presents the ablation study findings for our proposed SRP model across three benchmark datasets. We conduct ablation studies from the following three perspectives.

First, omitting the relation check part in Path Planning (\textit{- relation check}) significantly undermines performance of SRP, which underscores the necessity of validating relation correctness to generate reliable reasoning paths. 

Second, removing the sequence judge and path edit part in Reflection and Reasoning module (\textit{- self-reflection}) causes a notable drop on CWQ, revealing effects of reflection mechanisms for judging retrieved results and refining reasoning path. 

Third, eliminating the Reference Searching module (\textit{- reference}) yields a pronounced performance drop, highlighting the importance of reference; for instance, scores on WebQSP decline from 78.5\% to 75.7\%, demonstrating the benefits of leveraging external information for guidance. Furthermore, replacing the searched references with random-sampled references (\emph{random reference}) also reduces the result from 71.2\% to 70.9\% on GrailQA and from 78.5\% to 77.0\% on WebQSP, implying that carefully searched references are superior to arbitrary references.  

\begin{figure*}[t]
    \centering
    \includegraphics[width=1\linewidth]{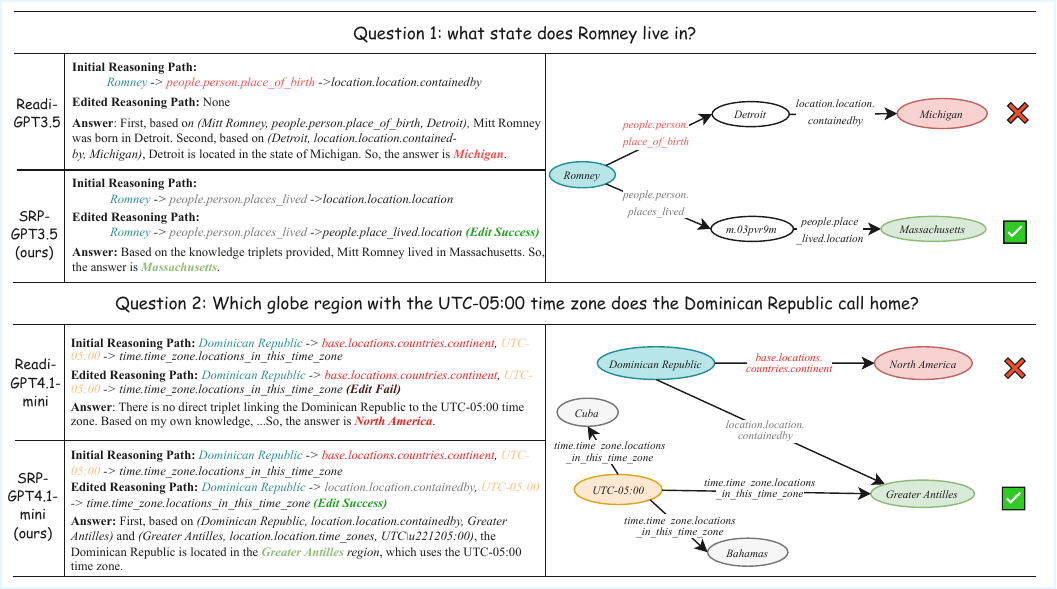}
    \caption{Case study of Readi and SRP. Question 1 comes from WebQSP, and question 2 comes from CWQ.}
    \label{fig:case}
\end{figure*}

\subsection{Reliability Study}
To further substantiate the reliability of our proposed SRP framework, we evaluate both the searching success rate and the reliable answering rate using the WebQSP, CWQ, and GrailQA datasets. The searching success rate denotes how efficiently each approach retrieves answer of question from the KG, while the reliable answering rate reflects defined as the proportion of correct answers supported by factual triples from the KG, thereby capturing the overall reliability of each system.

Table~\ref{tab:table_4} presents the searching success rate for of Readi-GPT3.5, Readi-GPT4.1-mini, SRP-GPT3.5, and SRP-GPT4.1-mini. The data illustrate that SRP's performance of retrieval valuable knowledge from KGs substantially exceeds Readi on three datasets. Even when Readi uses GPT4.1-mini, Readi's searching success rate is still lower than SRP-GPT3.5 on WebQSP, CWQ, and GrailQA, which demonstrates that the reasoning paths generated by SRP through reliable planning and reflection are more effective in retrieving the knowledge needed to answer the question.

As illustrated in Figure~\ref{fig:data}, we analyze each model’s reliable answering rate. Overall, SRP consistently outperforms Readi across three datasets. For instance, on WebQSP, SRP-GPT4.1-mini achieves a reliable answering rate of 92.2\%, significantly higher than the 70.5\% reached by Readi-GPT4.1-mini. The gap remains evident on CWQ and persists in GrailQA, where SRP maintains over 95\% reliability. These findings indicate that SRP’s correct answers are more frequently grounded in triples from KGs, reflecting strong factual support and high explanatory credibility.

\subsection{Case Study}
Figure~\ref{fig:case} illustrates two typical example, where Question 1 comes from WebQSP and Question 2 comes from CWQ. To further highlight the reliability of the SRP in generating and correcting reasoning paths, we conducted a case study on the WebQSP and CWQ datasets. As shown in Figure~\ref{fig:case}, we present the initial reasoning paths, edited reasoning paths, and the answers inferred based on retrieval results of SRP and Readi. The retrieval results are presented in the form of node graphs. GPT3.5 and GPT4.1-mini are utilized in question 1 and in question 2 respectively.

In Question 1, with reliable planning, SRP selects the correct initial relation (\textit{people.person\allowbreak.place\_of\_birth}), thereby establishing a solid foundation for generating correct reasoning path. This deliberate approach ultimately guides SRP to identify “Michigan” as the most plausible answer. Conversely, Readi initiates its reasoning path with an inappropriate relation (\textit{people.person.places\_lived}), which leads the reasoning path of Readi to deviate from the correct answer.

In Question 2, although SRP’s initial reasoning path is not flawless, by reflection on retrieved information, SRP swiftly edits the path with correct relation ``\textit{location.location.containedby}''. In contrast, Readi fails to realise that relation ``\textit{base.locations.countries.continent}'' was inconsistent with the problem due to lack of reflection on triplets sequence. The capacity for iterative refinement illustrates SRP’s resilience and adaptability, underscoring the importance of reflection in knowledge-based QA.

These cases demonstrate the reliability of the reasoning paths generated by SRP through the reliable planning and reflection. Through these reasoning paths, SRP can obtain more fact knowledge relevant to answering questions from the KG, thereby enhancing the reliability of reasoning.

\section{Conclusion}

In this paper, we propose Self-Reflective Planning (SRP) framework to address the limitations of LLMs in generating reliable reasoning path due to implicit knowledge reliance. To strengthen the reliability of its reasoning, SRP searches references for guidance, enforcing systematic relation-checking to ensure initial reasoning path alignment with structures of KGs, and repeatedly refining these reasoning path through self-reflection mechanism that preserves factual consistency. Extensive experiments on three challenging QA benchmarks demonstrate SRP's superior performance, achieving state-of-the-art results. We hope our work will motivate future studies about KG-enhanced LLM question answering.

\section*{Limitations}
While SRP shows promising results demonstrates that SRP can retrieve valuable knowledge with reliable reasoning path, two limitations still exists in SRP which should be resolved in future work.

First, the iterative reflection process increases computational costs compared to single-pass methods. This is because the reflection mechanism iteratively utilizes LLM to judge and edit for refining reasoning path, increasing the number of LLM calls. We hope to explore reliable and more efficient mechanisms to reduce computational costs in future research.

Second, performance of Reference Searching partially depends on the quality of the reference, which could be impacted by domain shifts. Because we use data from training set, this limitation doesn't influence the performance of SRP. Future work will explore dynamic reference adaptation and lightweight verification mechanisms to enhance the robustness of Reference Searching.

\bibliography{anthology}
\bibliographystyle{acl_natbib}

\appendix
\section{Datasets}
\label{appendix:data}
We adopt three widely adopted multi-hop knowledge graph question answering (KGQA) datasets.

\begin{itemize}
    \item WebQuestionsSP (WebQSP) \citep{yih-etal-2016-value} is a popular KGQA dataset derived from the WebQuestions dataset \citep{berant2013semantic}, designed for complex question answering over Freebase \citep{bollacker2008freebase}. It contains questions annotated with SPARQL queries, enabling research on semantic parsing and knowledge base question answering. 
    
    \item Complex WebQuestions(CWQ) \citep{talmor-berant-2018-web} extends the original WebQSP dataset by introducing compositional questions that require multi-hop reasoning. It consists of natural language questions that are more complex and involve multiple relations or constraints.
    
    \item GrailQA \citep{gu2021beyond} is a large-scale KGQA dataset built on Freebase \citep{bollacker2008freebase}. It evaluate models' generalization abilities across three levels: I.I.D., compositional, and zero-shot. 
\end{itemize}

\section{Baselines}
\label{appendix:baseline}
We comepare SRP with baselines grouping into 3 categories: 1) Fine-tuned model method; 2) LLM only methods; 3) Prompting LLM methods. The details of each baseline are describe as follows.

\paragraph{Fine-tuned model methods:}

\begin{itemize}
    \item TransferNet \citep{shi-etal-2021-transfernet} utilizes a transparent framework to transfer knowledge across domains by aligning feature distributions for improved domain adaptation.
    \item TIARA \citep{shu-etal-2022-tiara} is a framework that leverages BERT for schema item retrieval and T5 for plan generation, employing constrained decoding to ensure grammatical correctness.
    \item SR+NSM+E2E \citep{zhang-etal-2022-subgraph} trains an encoder to retrieve relevant relations and constructs reasoning paths based on the retrieved relations.
    \item Flexkbqa \citep{li2024flexkbqa} is a flexible KGQA framework that uses LLMs to adapt to different knowledge graphs and query languages with minimal annotated data.
\end{itemize}

\paragraph{LLM only methods:}

\begin{itemize}
    \item GPT3.5 \citep{openai_gpt35_turbo} is an advanced AI language model developed by OpenAI, capable of understanding and generating human-like text across a wide range of tasks.
    \item GPT4.1-mini \citep{openai_gpt4.1_mini} is a streamlined version of the GPT-4.1 model, optimized for efficient natural language processing with reduced computational demands while maintaining core performance capabilities.
\end{itemize}

\paragraph{Prompting LLM methods:}

\begin{itemize}
    \item KB-BINDER \citep{li-etal-2023-shot} is designed to address the heterogeneity of items across different KGs, facilitating few-shot in-context learning for KGQA tasks.
    \item StructGPT \citep{jiang2023structgpt} utilizes an interface for KG data to enable finite knowledge access and filtering, leveraging a LLM to repeatedly infer answers or subsequent planning.
    \item ToG \citep{sun2023think} is a training-free framework that incorporates LLMs with KGs through iterative beam search, enhancing deep reasoning capabilities, ensuring knowledge traceability and correctability, and enabling flexible, cost-effective deployment across diverse models and datasets.
    \item Readi \citep{cheng2024call} is a novel framework that enables LLMs to efficiently and faithfully reason over structured environments by initially generating a reasoning path, instantiating it on the environment, and invoking targeted editing only when necessary.
\end{itemize}

Table~\ref{tab:table_tog} presents performance of ToG-GPT3.5, ToG-GPT4, SRP-GPT3.5 and SRP-GPT4.1-mini on three datasets, where ToG-GPT4 means ToG motivated by GPT-4. Results of ToG-GPT3.5 and ToG-GPT4 are cited from \citep{chen2024plan}. Because of the lack of experiment data on ToG motivated by GPT4.1-mini and the difference of parameters size between GPT4 and GPT4.1-mini, we doesn't present results of ToG-GPT4 and compare ToG-GPT4 with SRP-GPT4.1-mini in table \ref{tab:table_2}. As shown in table~\ref{tab:table_tog}, SRP performs better than ToG when motivated by GPT-3.5. Although the performance of SRP-GPT4.1-mini is slightly inferior to that of ToG-GPT4. in GrailQA, SRP-GPT4.1-mini is better than ToG-GPT4 in WebQSP and CWQ. More importantly, the performance gap on GrailQA can be reasonably attributed to the fact that GPT-4.1-mini has significantly fewer parameters than GPT-4, which may limit its capacity in complex multi-hop reasoning tasks. Despite this, SRP-GPT4.1-mini still demonstrates strong generalization and competitive performance, indicating the robustness of our SRP framework even under lighter backbone models.

\begin{table*}[htbp]
  \centering
    \begin{tabular}{lcccccc}
    \toprule
    \multirow{2}[4]{*}{Method} & \multirow{2}[4]{*}{WebQSP} & \multirow{2}[4]{*}{CWQ} & \multicolumn{4}{c}{GrailQA} \\
\cmidrule{4-7}          &       &       & overall & I.I.D. & Compositional & Zero-shot \\
    \midrule
    ToG-GPT3.5 & 76.2  & 57.1  & 68.7  & 70.1  & 56.1  & 72.7 \\
    SRP-GPT3.5 & \textbf{78.6}  & \textbf{58.7}  & \textbf{71.2}  & \textbf{68.3}  & \textbf{58.6}  & \textbf{76.9} \\
    \midrule
    ToG-GPT4 & 82.6  & 67.6  & \textbf{81.4}  & \textbf{79.4}  & \textbf{67.3}  & \textbf{86.5} \\    
    SRP-GPT4.1-mini & \textbf{83.6}  & \textbf{69.7}  & 78.8  & 75.8  & 62.6  & 85.8 \\
    \bottomrule
    \end{tabular}%
  \caption{Performance of ToG-GPT3.5, ToG-GPT4, SRP-GPT3.5 and SRP-GPT4.1-mini on three datasets.}
  \label{tab:table_tog}%
\end{table*}%

\section{Prompts}
\label{appendix:prompts}
We provide prompts of SRP in this section. Prompts of relation check, path generation, sequence judge, path edit and answering are in Table \ref{tab:table_5}, Table \ref{tab:table_6}, Table \ref{tab:table_7}, Table \ref{tab:table_8} and Table \ref{tab:table_9}. Noted that the demonstration of prompts of relation check is from \citep{sun2023think}, and prompts of answering is following \citep{cheng2024call}.

For number of few-shot demonstration, we utilize 3 shots in WebQSP and 4 shots in CWQ and GrailQA for path generation. In path edit, we utilized 5 shots in WebQSP and CWQ and 4 shots in GrailQA. For relation check, sequence judge and answering, we utilizes 1 shots, 2 shots and 5 shots in each dataset respectively. When test Readi in GrailQA, we utilized 6 shots prompts for reasoning path generation and 4 shots demonstration for reasoning path edit. 

\begin{table*}[htbp]
\centering
\begin{tabular}{@{}p{0.95\textwidth}@{}}
\toprule
\textbf{Instruction}\\
Please retrieve 3 relations (separated by semicolon) that contribute to the question and rate their contribution on a scale from 0 to 1 (the sum of the scores of 3 relations is 1). Note: (1) please refer to the 4 examples of relation paths to give your score, if some example are similar to the question and the first relation of its relation path appears in candidate relation, give this relation a good rate; (2) please output relation and score in the format of ('relation', score).\\
\midrule
\textbf{Reference}\\
Here are 4 examples of questions and associated relation paths which connect to correct answer of question:\\
Question: \{reference question\}\\
Relation Path: \{reference reasoning path\}\\
......\\
\midrule
\textbf{Demonstration Example}\\
Question: Name the president of the country whose main spoken language was Brahui in 1980?\\
Topic Entity: Brahui Language\\
Candidate Relations: language.human\_language.main\_country; language.human\_language.language\allowbreak\_family; language.human\_language.iso\_639\_3\_code; base.rosetta.languoid.parent; language.human\\\_language.writing\_system; base.rosetta.languoid.languoid\_class; language.human\_language\allowbreak.countries\_spoken\_in; kg.object\_profile.prominent\_type; base.rosetta.languoid.document; base\allowbreak.ontologies.ontology\_instance.equivalent\_instances; base.rosetta.languoid.local\_name; language\allowbreak.human\_language.region
Answer:\\
1. ('language.human\_language.main\_country', 0.4): This relation is highly relevant as it directly relates to the country whose president is being asked for, and the main country where Brahui language is spoken in 1980.\\
2. ('language.human\_language.countries\_spoken\_in', 0.3): This relation is also relevant as it provides information on the countries where Brahui language is spoken, which could help narrow down the search for the president.\\
3. ('base.rosetta.languoid.parent', 0.2): This relation is less relevant but still provides some context on the language family to which Brahui belongs, which could be useful in understanding the linguistic and cultural background of the country in question.\\
\bottomrule
\end{tabular}
\caption{Prompts of relation check.}
\label{tab:table_5}
\end{table*}

\begin{table*}[htbp]
\centering
\begin{tabular}{@{}p{0.95\textwidth}@{}}
\toprule
\textbf{Instruction}\\
You are tasked with generating relation paths to help searching for answers in Freebase based on given question. I will provide you with:\\
1. A question.\\
2. One or more topic entity that is central to the question.\\
3. A set of valuable relations associated with the topic entity.\\

Your goal is to generate relation paths that start with the topic entity and follow a sequence of relations to help answer the question. \\
\midrule
\textbf{Demonstration Example}\\
Question: who played princess leia in star wars movies?\\
Topic Entity: princess leia\\
Valuable Relations: \{``princess leia'': ['film.film\_character.portrayed\_in\_films', 'tv.tv\_character\allowbreak.appeared\_in\_tv\_program', 'film.film\_character.movie', 'movie.movie\_character.movie']\}\\
Thought: Firstly, the path should cover the movies portrying princess leia. Secondly, the path should cover the actors in that movie.\\
Path: \{\\
\quad``princess leia'':[\\
\quad\quad``princess leia -> film.film\_character.portrayed\_in\_films -> film.performance.actor'', \\
\quad\quad``princess leia -> movie.movie\_character.movie -> film.actor.actor'' \\
\quad]\\
\}\\
\bottomrule
\end{tabular}
\caption{Prompts of path generation.}
\label{tab:table_6}
\end{table*}

\begin{table*}[htbp]
\centering
\begin{tabular}{@{}p{0.95\textwidth}@{}}
\toprule
\textbf{Instruction}\\
Here are some triplet sequences [(h\_0, r\_0, t\_0), ..., (h\_n, r\_n, t\_n)] that may contain information helpful for solving the problem. Please analyze the following triplet sequences and retain the subsequences within each triplet sequence that are useful for answering the question, while removing the subsequences that are not helpful. Please first output your Thinking Process, then output the retained parts of each triplet sequence. If you believe the answer to the question appears at the end of the triplet sequence (i.e., the answer is the tail entity t\_n of the last triplet), directly return this sequence. If you think that the entire triplet sequence, except for the head entity h\_0 of the first triplet, is unrelated to the question, return an empty list [].\\ 
Note: (1) The retained part of the triplet sequence should be a continuous subsequence, and the removed part should also be a continuous subsequence; you cannot return non-continuous triples from the original sequence. (2) If it is possible to retain, the retained part should include at least the first triplet of the sequence. (3) The format of the output triplet sequence should be the same as the input triplet sequence. (4) If you believe the answer to the question appears in the triplet sequences, please give ``<HAVE\_ANSWER>'' in the end of your Thinking Process. If you do not believe the answer to the question appears in the triplet sequences, please give ``<NO\_ANSWER>'' in the end of your Thinking Process.\\
\midrule
\textbf{Reference}\\
Here are 4 examples of some questions, associated relation and answer of question.\\
Question: \{reference question\}\\
Relation Path: \{reference reasoning path\}\\
Answer: \{reference answer\}\\
......\\
\midrule
\textbf{Demonstration Example}\\
Question: where is aviano air force base located?\\
Triplet sequences:\\
1. [(``Aviano Air Base'', ``location.location.containedby'', ``Italy'')]\\
2. [(``Aviano Air Base'', ``aviation.airport.serves'', ``Aviano'')]\\
Thinking Process: First, based on the triplet (``Aviano Air Base'', ``location.location.containedby'', ``Italy''), I can answer the question. So, I think these triplet sequences have enough information to answer the question. <HAVE\_ANSWER>\\
Retained sequences:\\
1. [(``Aviano Air Base'', ``location.location.containedby'', ``Italy'')]\\
2. []\\
\bottomrule
\end{tabular}
\caption{Prompts of sequence judge.}
\label{tab:table_7}
\end{table*}

\begin{table*}[htbp]
\centering
\begin{tabular}{@{}p{0.95\textwidth}@{}}
\toprule
\textbf{Instruction}\\
Task: Given an Inital Path and some feedback information of a Question, please correct the Inital Path.\\
Note:\\
(1)When you receive Error Message, please edit the path based on Instantiate Paths. For example, if the Error Message is ``relation XXX not instantiated'', you should modify this relation with candidate relation; if the Error Message is ``<cvt></cvt> in the end'', you should add a candidate relation to a Instantiate Path which you think is relevant to question; if the Error Message is ``Current Information is not enough'', please analysis Instantiate Paths and Candidate Relations, then generate a new path which is more relevant to question; (2) please refer to the 4 examples of relation paths to correct the Inital Path; (3) Avoid generating Final Path that are the same as the Initial Path.\\
\midrule
\textbf{Reference}\\
Question: \{reference question\}\\
Relation Path: \{reference reasoning path\}\\
......\\
\midrule
\textbf{Demonstration Example}\\
Question: What major religion in the UK has a place of worship named St. Mary's Cathedral, Batticaloa?\\
Initial Path: United Kingdom -> location.location.religions -> place.religion.major\_religions\\
>>>> Error Message\\
1. <cvt></cvt> in the end. \\
2. relation ``place.religion.major\_religions'' not instantiated.\\
>>>> Instantiation Context\\
Instantiate Paths: United Kingdom -> location.location.contains -> Heaton railway station\\
United Kingdom -> location.statistical\_region.religions -> <cvt></cvt>\\
United Kingdom -> location.location.contains -> Bakersfield, Nottingham\\
United Kingdom -> location.location.contains -> Knockloughrim\\
United Kingdom -> location.location.contains -> Oakenshaw\\
Candidate Relations: \{'United Kingdom -> location.statistical\_region.religions': ['location.religion\\\_percentage.date', 'location.religion\_percentage.percentage', 'location.religion\_percentage\allowbreak.religion'], 'United Kingdom -> location.location.contains': ['location.location.containedby', 'location.location.geolocation', 'type.object.type']\}\\
>>>> Corrected Path\\
Goal: The Initial Path starts from United Kingdom, which should cover the major religion in United Kingdom.\\
Thought: In Instantiate Paths, I find that United Kingdom has some religions, described by a cvt node.
In candidates, I find ``location.religion\_percentage.religion'' most relevant to major religions.\\
Final Path: United Kingdom -> location.statistical\_region.religions -> location.religion\allowbreak\_percentage.religion\\
\bottomrule
\end{tabular}
\caption{Prompts of path edit.}
\label{tab:table_8}
\end{table*}

\begin{table*}[htbp]
\centering
\begin{tabular}{@{}p{0.95\textwidth}@{}}
\toprule
\textbf{Instruction}\\
Given a question and the associated retrieved knowledge graph triplets (entity, relation, entity), you are asked to answer the question with these triplets. When you answer the question, please first give your answer with your own knowledge, then give your answer with knowledge from retrieved knowledge graph triplets. If the given knowledge triples is not enough or missing, you can use your own knowledge. Use \{\} to enclose the answer! Please think step by step.\\
\midrule
\textbf{Demostration Example}
Q: Find the person who said ``Taste cannot be controlled by law'', where did this person die?
Knowledge Triplets: (Taste cannot be controlled by law., media\_common.quotation.author, Thomas Jefferson)
A: First, based on (Taste cannot be controlled by law., media\_common.quotation.author, Thomas Jefferson), the person who said ``Taste cannot be controlled by law'' is Thomas Jefferson. Second, no Triplet provided can answer where Thomas Jefferson's dead, however, based on my owned knowledge, Thomas Jefferson died in Charlottesville. So, the answer is { Charlottesville }.\\
\bottomrule
\end{tabular}
\caption{Prompts of answering.}
\label{tab:table_9}
\end{table*}

\end{document}